\documentclass{article}

\usepackage[utf8]{inputenc} 
\usepackage[T1]{fontenc}    
\usepackage{url}            
\usepackage{booktabs}       
\usepackage{amsfonts}       
\usepackage{nicefrac}       
\usepackage{amsmath}
\usepackage{latexsym}
\usepackage{amssymb}
\usepackage{amsthm} 
\usepackage{graphics,color}
\usepackage{microtype}
\usepackage{graphicx}
\usepackage{subfig}
\usepackage{booktabs} 

\newtheorem{theorem}{Theorem}
\newtheorem{lemma}{Lemma}

\newtheorem{corollary}{Corollary}
\newtheorem{assumption}{Assumption}

\theoremstyle{definition}

\newtheorem{remarks}{Remark}
\newtheorem{example}{Example}

\def\mc{\mathcal}
\usepackage{soul}
\def\mbf{\mathbf}
\def\mbb{\mathbb}
\def\mbs{\boldsymbol}
\def\P{\mathsf{P}}
\def\expe{\mathbb{E}}   
\def\argmin{\mathop{\rm argmin}}
\def\argmax{\mathop{\rm argmax}}

\newcommand{\dist}[3]{l_{#1}\left(#2; #3\right)}
\newcommand{\samp}[2]{Y_{#1}^{(#2)}}

\newcommand{\bel}[2]{b_{#1}^{(#2)}}
\newcommand{\est}[2]{\rho_{#1}^{(#2)}}

\usepackage[accepted]{icml2019}

\usepackage{etoolbox}
\newcommand{\algorithmicdoinparallel}{\textbf{do in parallel}}
\makeatletter
\AtBeginEnvironment{algorithmic}{%
  \newcommand{\FORP}[2][default]{\ALC@it\algorithmicfor\ #2\ %
    \algorithmicdoinparallel\ALC@com{#1}\begin{ALC@for}}%
}
\makeatother

\begin{document}
\allowdisplaybreaks
\twocolumn[
\icmltitle{Peer-to-Peer Federated Learning on Graphs}



\icmlsetsymbol{equal}{*}

\begin{icmlauthorlist}
\icmlauthor{Anusha Lalitha}{to}
\icmlauthor{Osman Cihan Kilinc}{to}
\icmlauthor{Tara Javidi}{to}
\icmlauthor{Farinaz Koushanfar
}{to}
\icmlaffiliation{to}{Department of Electrical Computer Engineering, University of California, San Diego, USA}
\end{icmlauthorlist}

\icmlaffiliation{to}{Department of Electrical Computer Engineering, University of California, San Diego, USA}

\icmlcorrespondingauthor{Anusha Lalitha}{alalitha@eng.ucsd.edu}

\icmlkeywords{Machine Learning, ICML}

\vskip 0.3in
]


\printAffiliationsAndNotice{} 

\begin{abstract}
We consider the problem of training a machine learning model over a network of nodes in a fully decentralized framework. The nodes take a Bayesian-like approach via the introduction of a belief over the model parameter space. We propose a distributed learning algorithm in which nodes update their belief by judicially aggregating information from their local observational data with the model of their one-hop neighbors to collectively learn a model that best fits the observations over the entire network. Our algorithm generalizes the prior work on federated learning. Furthermore, we obtain theoretical guarantee (upper bounds) that the probability of error and true risk are both small for every node in the network.  We specialize our framework to two practically relevant problems of linear regression and the training of Deep Neural Networks (DNNs). 
\end{abstract}

\vspace{-0.3cm}
\section{Introduction}

Mobile computing devices have seen a rapid increase in their computational power as well as storage capacity. 
Aided by this increased computational power and abundance of data, as well as due to privacy and security concerns, there is a growing trend towards training machine learning models cooperatively over networks of such devices using only local training data. 
The field of \emph{Federated learning} initiated by \citet{mcmahan2017communication} and \citet{konevcny2016federated} considers the problem of learning a centralized model based on private training data of a number of nodes.
More specifically, this framework is characterized by a  possibly large number of decentralized nodes which are (i) connected to a centralized server and (ii) have access to only local training data possibly correlated across the network. It is also assumed that communications between the nodes and the central server incur large costs. \citet{mcmahan2017communication} proposed the \emph{federated optimization} algorithm in which  the central server randomly selects a fraction of the nodes in each round, shares the current global model with them, and then averages the updated models sent back to the server by the selected nodes. \citet{mcmahan2017communication} and \citet{konevcny2016federated} also provided experimental results with good accuracy using both convolutional and recurrent neural networks while reducing the communication costs. 

\vspace{-0.1cm}
This work generalizes the model and the framework of federated learning framework of~\citet{mcmahan2017communication} in the following important directions. Conceptually our contributions are as follows:\vspace{-0.3cm}
\begin{itemize}\itemsep0em
    \item \textbf{Fully Decentralized Framework:} We do not require a centralized location where all the training data is collected or a centralized controller to maintain a global model over the network by aggregating information from all the nodes. Instead, in our setting, nodes are distributed over a network/graph where they only communicate with their one-hop neighbors. Hence, our problem formulation does away with the need of having a centralized controller. 
    \vspace{-0.1cm}
    \item \textbf{Localized Data:} We allow the training data available to an individual node to be insufficient for learning the shared global model. In other words, the nodes must collaborate with their next hop neighbors to learn the optimal model even though for privacy concerns, nodes do not share their raw training data with the neighbors. 
\end{itemize}
\vspace{-0.3cm}
To motivate our work and underline our contributions, consider the following simple toy example.

\begin{example}[Distributed Linear Regression]
\label{ex:dist_linear_regression}
Let $d \geq 2$ and $\Theta = \mbb{R}^{d+1}$. For $\mbs{\theta}= [\theta_0, \theta_1, \ldots, \theta_d]^T \in \Theta$, $\mbf{x} \in \mbb{R}^d$, define 
\vspace{-0.3cm}
\begin{align*}
 f_{\mbs{\theta}}(\mbf{x}) := \theta_0 + \sum_{i=1}^{d} \theta_i x_{i} =  \langle \mbs{\theta},  [1, \mbf{x}^T]^T \rangle.   
\end{align*}
The label variable $y \in \mbb{R}$ is given by a deterministic function $f_{\mbs{\theta}}(\mbf{x})$ with additive Gaussian noise $\eta \sim N(0, \alpha^2)$ so that\vspace{-0.2cm}
\begin{align}
\label{eq:regression}
y = f_{\mbs{\theta}}(\mbf{x}) + \eta.
\end{align}
Consider a network of $N$ nodes. Consider the realizable setting where there exists a $\mbs{\theta}^{\ast} \in \Theta$ which generates the labels $y \in \mbb{R}$ as given by~\eqref{eq:regression}. Fix some $0<m<d$ and let $\mc{X}_1 = \left\{ \left. \begin{bmatrix} \mbf{x} \\ 0\end{bmatrix} \right| \mbf{x} \in \mbb{R}^{m} \right\}$ and $\mc{X}_2 =\left\{ \left. \begin{bmatrix} 0 \\ \mbf{x}\end{bmatrix} \right| \mbf{x} \in \mbb{R}^{d-m} \right\}$. 
Now let us assume that each node is 
one of two types: type-1 nodes make observations corresponding to points in $\mc{X}_1$ and type-2 nodes in $\mc{X}_2$. In the absence of communication between type-1 ad type-2 nodes (dictated by the graph structure), the deficiency of the local data prevents type-1 nodes to disambiguate the set
$
   \overline{\Theta}_1 = \left\{ \mbs{\theta} \in \Theta \mid \mbs{\theta}_{(0:m)} = \mbs{\theta}^{\ast}_{(0:m)}\right\}.
$
\end{example}

\vspace{-0.3cm}
Our fully decentralized federated learning, when specialized to this regression example, provides an information exchange rule which, despite the generality of the graph and the deficiency of the local observations, result in each node eventually learning the true parameter $\mbs{\theta}^*$.
Our theoretical contributions are as follows:\vspace{-0.3cm}
\begin{itemize}
    \item 
    Mathematically, we pose the problem of federated machine learning as a special case of the problem of social learning on a graph. Social learning on a graph has long been studied in statistics, economics, and operations research and encompasses canonical problems of consensus ~\citet{DeGroot:74consensus},  belief propagation~\cite{OlfatiSaberFES:05belief}, and distributed hypothesis testing~\cite{Jadbabaie13informationheterogeneity,7172262, 7349151, 8359193}.  To the best of our knowledge, our proposed formulation is the first to make this connection, allowing for the application of a gamut of statistical tools from social learning in the context of federated learning. 
    \vspace{-0.2cm}
    \item
    In particular, borrowing from~\cite{7172262, 7349151, 8359193}, we propose a peer-to-peer social learning scheme where the nodes take a Bayesian-like approach via the introduction of a belief over a parameter space characterizing the unknown global (underlying) model.  Fully decentralized learning of the global (underlying) model is then achieved via a two step procedure. First, each node updates its local belief according to a Bayesian inference step (posterior update) based on the node's local data. This step is, then, followed by a consensus step of aggregating information from the one-hop neighbors. 
    \vspace{-0.2cm}
    \item
    Under mild constraints on the network connectivity and global learnability of the network, we provide high probability guarantees on the number of training samples required so that the nodes each learn the globally optimal model that best fits the samples across the network.

\end{itemize}
\vspace{-0.2cm}
Empirically, we validate our theoretical framework in two specific, yet canonical, linear and non-linear machine learning problems: linear regression and training of deep neural network (DNN). Our proposed social learning algorithm and its theoretical analysis rely on a local Bayesian posterior update, which in most practically relevant applications such as training of DNNs turn out to be computationally intractable. To overcome this, we employ variational inference (VI) \citep[Chapter~3]{gal2016uncertainty} techniques which replace the Bayesian modelling marginalization with optimization. Our experiments on a network of two nodes cooperatively training a shared DNN show that fully decentralized federated learning can be done with little to no drop in accuracy relative to a central node with access to all the training data.



\noindent \textbf{Notation}:  We use boldface for vectors and denote the $i$-th element of vector $\mathbf{v}$ by $v_{i}$. For any two vectors $\mbf{x}$ and $\mbf{y}$, let $\langle \mbf{x}, \mbf{y} \rangle$ denote the dot product between $\mbf{x}$ and $\mbf{y}$. For any vector $\mbf{x}$, let diag$(\mbf{x})$ denote the diagonal matrix with diagonal elements given by $\mbf{x}$. We let $[n]$ denote $\{1, 2, \ldots, n\}$. $\mc{P}(A)$ denotes the set of all probability distributions on a set $A$ and $| A|$ denotes the number of elements in set $A$. Let $D_{\text{KL}}(P_{Z}||P_Z')$ the Kullback--Leibler (KL) divergence between two probability distributions $P_Z, P_Z' \in \mc{P}(\mc{Z})$.



\vspace{-0.2cm}
\section{Problem Setup}
 In this section, we formally describe the label generation model at each node, the communication graph, and a criterion for successful learning over the network.
\vspace{-0.2cm}
\subsection{The Model}
Consider a group of $N$ individual nodes. 
Each node $i \in [N]$ has access to a dataset $\mc{D}_i$ consisting of $n$ instance-label pairs, $\left(X_i^{(k)}, Y_i^{(k)}\right)$ where $k \in [n]$. 
Each instance $X_i^{(k)} \in \mc{X}_i \subseteq \mc{X}$, where $\mc{X}_i$ denotes the local instance space of node $i$ and $\mc{X}$ denotes a global instance space which satisfies $\mc{X} \subseteq \cup_{i=1}^N \mc{X}_i$. 
Similarly, let $\mc{Y}$ denote the set of all possible labels over all the nodes. Some examples include, $\mc{Y} = \mbb{R}$ for regression and $\mc{Y} = \{0,1\}$ for binary classification. 
The samples $\left\{X_i^{(1)}, X_i^{(2)}, \ldots, X_i^{(n)}\right\}$ are independent and identically distributed (i.i.d), and are generated according to a distribution $\P_i \in \mc{P}(\mc{X}_i)$.  
We view the model generating the labels for each node $i$ as a probabilistic model with a distribution $f_i(y| x)$, $\forall\,y\in \mc{Y}, \forall x \in \mc{X}$.

Consider a finite parameter set $\Theta$ with $M$ points. We assume that each node has access to a set of \textit{local likelihood functions} of the labels $\{\dist{i}{y}{\theta, x}:  y \in \mc{Y}, \, \theta \in \Theta, \, x \in \mc{X}_i \}$, where $\dist{i}{y}{\theta, x}$ denotes the local likelihood function of label $y$, given $\theta$ is the true parameter, and instance $x$ was observed. 
For each $i$, define 
\begin{align*}
\overline{\Theta}_i := \argmin_{\theta \in \Theta}\expe_{\P_i}\left[D_{\text{KL}}\left(f_i(\cdot|X_i)||l_i(\cdot| \theta, X_i)\right)\right].    
\end{align*}
Furthermore, define $\Theta^{\ast} := \cap_{i=1}^N \overline{\Theta}_i$. 
In this work, we are interested in the case where $\Theta^{\ast} \neq \phi$. Then we say any parameter $\theta^{\ast} \in \Theta^{\ast}$ is \textit{globally learnable}. 

\begin{assumption}
\label{assump:global_learnability}
There exists a parameter $\theta^{\ast} \in \Theta$ that is \textit{globally learnable}, i.e, $\cap_{i=1}^N \overline{\Theta}_i \neq \phi$. 
\end{assumption}
\vspace{-0.2cm}
Note that, if the local input space $\mc{X}_i$ is such that $|\overline{\Theta}_i| >1$, then learning a parameter $\theta^{\ast} \in \Theta^{\ast}$ is not possible using the local dataset of node $i$. However, under Assumption~\ref{assump:global_learnability} the nodes can collaborate over a network to learn $\theta^{\ast} \in \Theta^{\ast}$.
\vspace{-0.3cm}

\subsection{The Communication Network}
We model the communication network between nodes via a directed graph with vertex set $[N]$. We define the neighborhood of node $i$, denoted by $\mathcal{N}(i)$, as the set of all nodes $j$ who have an edge going from $j$ to $i$. Furthermore, if node $j \in \mathcal{N}(i)$, it can exchange information with node $i$. The social interaction of the nodes is characterized by a stochastic matrix $W$. The weight $W_{ij} \in [0, 1]$ is strictly positive if and only if $j \in \mc{N}(i)$ and  $W_{ii} = 1 - \sum_{j = 1}^{N} W_{ij}$. The weight $W_{ij}$ denotes the confidence node $i$ has on the information it receives from node $j$. We make the following assumption that allows the information gathered at every node to be disseminated throughout the network.  


\begin{assumption}
The network is a strongly connected aperiodic graph. Hence, $W$ is aperiodic and irreducible.
\end{assumption}

\subsection{The Learning Criterion}

We say that an algorithm learns a global learnable parameter $\theta^{\ast} \in \Theta^{\ast}$ in a distribution manner across the network if the following holds: for any confidence parameter $\delta \in (0,1)$ we have \vspace{-0.2cm}
\begin{align*}
    \P\left(\exists \, i \in [N] \,\, \text{s.t.} \,\, \hat{\theta}^{(n)}_i \not \in \Theta^{\ast} \right) \leq  \delta,
\end{align*}
where $\hat{\theta}^{(n)}_i \in \Theta$ denotes the estimate of node $i$ after observing $n$ instance-label pairs. Our learning criteria requires every node in the network to agree on a parameter that best fits the dataset distributed over the entire network. 

\section{Peer-to-peer Federated Learning Algorithm}
\label{sec:algorithm}

In this section, we consider a learning algorithm that generates the information that nodes exchange with each other and dictates the merging of all the information gathered at each node. We employ the distributed hypothesis testing algorithm considered by~\citet{7172262, 7349151, 8359193} to cooperatively learn the model over the network. 
At every instant $k$ each node $i$ maintains a private belief vector $\mbs{\est{i}{k}} \in \mc{P}(\Theta)$ and a public belief vector $\mbf{\bel{i}{k}} \in \mc{P}(\Theta)$.
At each instant $k \in [n]$, each node $i$ executes the algorithm described in Algorithm \ref{alg:algorithm}.

\begin{algorithm}[!htb]
   \caption{Peer-to-peer Federated Learning Algorithm}
   \label{alg:algorithm}
\begin{algorithmic}[1]
   \STATE{{\bfseries Inputs:} $\mbs{\est{i}{0}} \in \mc{P}(\Theta)$ with $\mbs{\est{i}{0}} > 0$ for all $i \in [N]$}
   \STATE{{\bfseries Outputs:} $\hat{\theta}^{(n)}_i$ for all $i \in [N]$}
   \FOR{instance $k = 1$ {\bfseries to} $n$}
   \FORP{node $i=1$ {\bfseries to} $N$}
   \STATE {Draw an i.i.d sample $X^{(k)}_i \sim\P_i$ and obtain a conditionally i.i.d sample label $\samp{i}{k} \sim f_i\left(\cdot|X^{(k)}_i\right)\P_i\left(X^{(k)}_i\right)$. }
   \STATE {
   	Perform a local Bayesian update on $\mbs{\est{i}{k-1}}$ to form the belief vector $\mbf{\bel{i}{k}}$ using the following rule.
	 For each $\theta \in \Theta$, 
		\begin{align}
		\bel{i}{k}(\theta) = \frac{ \dist{i}{\samp{i}{k}}{\theta, X^{(k)}_i} \est{i}{k-1}(\theta) }{ \sum_{\psi \in \Theta} \dist{i}{\samp{i}{k}}{\psi, X^{(k)}_i} \est{i}{k-1}(\psi) }.
		\label{eq:bayes}
		\end{align}
   }
   \STATE {
	Send $\mbf{\bel{i}{k}}$ to all nodes $j$ for which $i \in \mc{N}(j)$. Receive $\mbf{\bel{j}{k}}$ from neighbors $j \in \mc{N}(i)$.
	}
	\STATE {
	Update private belief by averaging the log beliefs received from neighbors , i.e., for each  $\theta \in \Theta$, 
		\begin{align}
		\est{i}{k}(\theta) = \frac{ \exp \left( \sum_{j = 1}^{N} W_{ij} \log \bel{j}{k}(\theta) \right)
			}{
			\sum_{\psi \in \Theta} \exp \left( \sum_{j = 1}^{N} W_{ij} \log \bel{j}{k}(\psi) \right)
			}.
		\label{eq:estimate}
		\end{align}
	}
	\STATE {
	Declare an estimate \vspace{-0.1cm}
	\begin{align*}
	    \hat{\theta}_i^{(k)} := \argmax_{\theta \in \Theta}  \est{i}{k}(\theta).
	\end{align*}
	}
	\ENDFOR
   \ENDFOR
	\end{algorithmic}
\end{algorithm}

%

\section{Analysis of Learning Algorithm}
To analyze our peer-to-peer learning algorithm, we will further require the following technical assumptions on the initial belief vectors and the likelihood functions. As demonstrated by our experiments in Section~\ref{sec:experiments} the performance of our learning algorithm is not affected when these assumptions are not satisfied.

\begin{assumption}
For all nodes $i \in [N]$, assume:
\begin{itemize}
    \item
    The prior beliefs $\est{i}{0} (\theta) > 0$ for all $\theta \in \Theta$.
    
    \item
    There exists an $\alpha > 0,\, L > 0$ such that $\alpha < l_i(y; \theta, x) < L$, for all $y \in \mc{Y}$, $\theta \in \Theta$ and $x \in \mc{X}_i$.
\end{itemize}
\end{assumption}

\begin{theorem}
\label{thm:error_bound}
Given a finite set $\Theta$ with $M$ parameters. Using the distributed learning algorithm described in Section~\ref{sec:algorithm}, for any given confidence parameter $\delta \in (0,1)$ we have 
\begin{align*}
    \P\left(\exists \, i \in [N] \,\, \text{s.t.} \,\, \hat{\theta}^{(n)}_i \not \in \Theta^{\ast} \right) \leq  \delta,
\end{align*}
when the number of training samples satisfies \vspace{-0.2cm}
\begin{align}
\label{eq:samples_lbd}
    n \geq \frac{16C\log \frac{N M}{\delta}}{ K(\Theta)^2(1-\lambda_{\text{max}}(W))},
\end{align}
\vspace{-0.2cm}where we define \vspace{-0.2cm}
\begin{align*}
K(\Theta):= \min_{\theta \in \Theta^{\ast}, \psi \in \Theta \setminus \Theta^{\ast}}\sum_{j = 1}^{N} v_j I_j(\theta, \psi),
\end{align*}
and  define \vspace{-0.2cm}
\begin{align*}
    I_j(\theta, \psi) 
    &:=  \expe_{\P_j}\left[ D_{\text{KL}}\left( f_j(\cdot| X_j)|| l_j(\cdot; \psi, X_j)\right) \right.
    \\
    &\left.\hspace{0.5cm}- D_{\text{KL}}\left( f_j(\cdot| X_j)|| l_j(\cdot; \theta, X_j)\right) \right],
\end{align*}
where $\mbf{v} = [v_1, v_2, \ldots, v_N]$ is the unique stationary distribution of $W$ with strictly positive components, $\lambda_{\text{max}}(W) = \max_{1\leq i \leq N-1}\lambda_i(W)$,  where $\lambda_i(W)$ denotes $i$-th eigenvalue of $W$ counted with algebraic multiplicity and $\lambda_0(W) = 1$, and $C := \left|\log \frac{L}{\alpha}\right|$. 
\end{theorem}

Proof of Theorem~\ref{thm:error_bound} is provided in Appendix~\ref{app:proof}.

\begin{remarks}
The lower bound on the number of training samples grows logarithmically in the number of nodes in the network and number of parameters to be distinguished. The lower bound also inversely depends on $K(\Theta)$ which dictates the smallest rate at which the nodes distinguish the parameter that best fits data from rest of the parameters across the network. Furthermore, it inversely depends on the rate which $W$ converges to its stationary distribution.
\end{remarks}


\vspace{-0.2cm}
\subsection{Upper Bound on True Risk}
We make the following realizability assumption.
\begin{assumption}
\label{assump:realizability}
There exists a parameter $\theta^{\ast} \in \Phi$, where parameter set $\Phi$ is a compact subset of $\mbb{R}^d$, such that $f_i(\cdot|x) = l_i(\cdot; \theta^{\ast}, x)$ almost everywhere for all nodes $i \in [N]$. Furthermore, we assume there exists a set $\Theta \subset \Phi$ of cardinality $M$ which is an $r$-covering of $\Phi$, i.e., $\Phi \subset \cup_{\theta \in \Theta}\mc{B}_r(\theta)$, where 
\begin{align}
\label{eq:r_covering}
&\mc{B}_r(\theta) :=
\nonumber
\\
&\left\{ \psi \in \Phi: \sum_{i=1}^N\frac{\expe_{\P_i}[D_{\text{KL}}(l_i(\cdot; \theta, X_i)||\dist{i}{\cdot}{\psi, X_i})]}{N} \leq r\right\}.
\end{align}
\end{assumption}
\vspace{-0.4cm}
In the above assumption, we assume that the model which generating labels across the network can be parametrized by a continuous parameter $\theta$ which belongs to a compact set $\Phi \subset \mbb{R}^d$. Furthermore, we assume there exists a quantization of $\Phi$ with quantization points in $\Theta$ such that $\Theta$ is an $r$-covering of $\Phi$ as specified by equation~\eqref{eq:r_covering}. From the definition of $\Theta^{\ast} = \cap_{i=1}^N\overline{\Theta}_i$, note that for any $\theta \in \Theta^{\ast}$ we have $\theta^{\ast} \in \mc{B}_r(\theta)$. Furthermore, using the peer-to-peer learning algorithm we learn $\hat{\theta}_i^{(n)} \in \Theta^{\ast} = \cap_{i=1}^N\overline{\Theta}_i$ with high probability. Hence, with high probability we have $\theta^{\ast} \in \mc{B}_r\left(\hat{\theta}^{(n)}_i\right)$ for all $i \in [N]$. Under Assumption~\ref{assump:realizability} we provide an upper bound on the true risk. Let $r_i(x, y)$ denote the risk function of node $i\in [N]$ associated with sample $(x,y) \in \mc{D}_i$. The expected risk at node $i$ when $\theta$ is the underlying parameter is given by $R_i(\theta) = \expe_{\P_i}\left[ \int_{\mc{Y}} r_i(x,y) \dist{i}{y}{\theta, x} dy\right]$. Now, using Theorem~\ref{thm:error_bound} we obtain the following bound on the average expected risk over the network as a corollary.
\begin{corollary}
Consider $|r_i(x,y)| \leq B$ for all $x \in \mc{X}_i$, $y \in \mc{Y}$, then under Assumption~\ref{assump:realizability} using the above algorithm with probability at least $1-\delta$ for the number of samples given by~\eqref{eq:samples_lbd} we have \vspace{-0.1cm}
    \begin{align*}
        &\frac{1}{N}\sum_{i=1}^N\left| R_i\left(\theta^{\ast}\right)-R_i\left(\hat{\theta}^{(n)}_i\right)\right|
        \\
        &\leq 
        \frac{B}{N}\sum_{i=1}^N\expe_{\P_i}\left[\int_{\mc{Y}} \left| \dist{i}{y}{\theta^{\ast}, x} - \dist{i}{y}{\hat{\theta}^{(n)}_i, x}\right|dy\right]
        \\
        &\overset{(a)}\leq 
        \frac{B}{2N}\sum_{i=1}^N\expe_{\P_i}\left[ \sqrt{D_{\text{KL}} \left( \dist{i}{y}{\theta^{\ast}, x} || \dist{i}{y}{\hat{\theta}^{(n)}_i, x} \right)}\right] 
        \\
        & \overset{(b)}\leq 
        \frac{B}{2} \sqrt{\frac{1}{N}\sum_{i=1}^N\expe_{\P_i}\left[ D_{\text{KL}} \left( \dist{i}{y}{\theta^{\ast}, x} || \dist{i}{y}{\hat{\theta}^{(n)}_i, x} \right)\right]}
        \\
        & \overset{(c)}\leq \frac{B\sqrt{r}}{2}, 
    \end{align*}
    where $(a)$ follows from Pinsker's inequality, $(b)$ from Jensen's inequality and $(c)$ follows from Theorem~1.

\end{corollary}


\vspace{-0.2cm}
\section{Experiments}

\label{sec:experiments}

In this section, we provide our experimental results. 

\subsection{Distributed Bayesian Linear Regression}
Consider a network of two nodes with one node each from type-1 and type-2 in Example~\ref{ex:dist_linear_regression} with $\Theta = \mbb{R}^3$ and $\mc{X} = \mbb{R}^2$ (hence, $d = 2$, $m = 1$) and $\mbs{\theta}^{\ast} = [-0.3, \, 0.5, \, 0.8]^T$. Let the edge weights be given by $\mbf{W} =\begin{bmatrix} 0.9 & 0.1\\ 0.6 & 0.4 \end{bmatrix}$. Suppose the observation noise is distributed as $\eta \sim \mc{N}(0, \alpha^2)$ where $\alpha = 0.8$. Training data $\mc{D}_{1}$ of node 1 consists of instance-label pairs for $[x_1 , 0]^T \in \mbb{R}^2$ where $x_1$ is sampled from $\text{Unif}[-1,1]$ and training data  $\mc{D}_{2}$ of node 2 consists of instance-label pairs for $[0, x_2]^T \in \mbb{R}^2$ where $x_2$ is sampled from $\text{Unif}[-1.5, 1.5]$. However, the test set consists of observations belonging to $\mbf{x} \in \mbb{R}^2$. We assume each node starts with a Gaussian prior over $\mbs{\theta}$ with zero mean $[0, 0 , 0]^T$ and covariance matrix given by $\text{diag}[0.5, 0.5, 0.5]$.

When there is no cooperation among the nodes, each node aims to learn the respective posterior distribution $\P(\mbs{\theta}| \mc{D}_{i})$, and make predictions on the test set using the predictive distribution $\P(y| \mc{D}_{i}) = \int \P(y| \mbs{\theta})\P(\mbs{\theta}| \mc{D}_{i}) d\mbs{\theta}$, where $i \in \{1, 2\}$. When there is cooperation among the nodes, we learn the posterior distribution on $\mbs{\theta}$ using Algorithm~\ref{alg:algorithm}. Note that since the nodes begin with a Gaussian prior on $\mbs{\theta}$, their beliefs after a Bayesian update remain Gaussian. Furthermore, suppose $\mbs{\bel{i}{k}} \sim \mc{N}(\mbs{\mu}_i, \mbs{\Sigma}_i)$ where $i \in \{0,1\}$, then equation~\eqref{eq:estimate} reduces to 
\begin{align}
\label{eq:merge_mu}
    \tilde{\mbs{\Sigma}}_i^{-1} &= \mbf{W}_{i1}\mbs{\Sigma}_1^{-1} + \mbf{W}_{i2}\mbs{\Sigma}_2^{-1},
    \\
    \tilde{\mbs{\mu}}_i &= \tilde{\mbs{\Sigma}}_i\left( \mbf{W}_{i1}\mbs{\Sigma}_1^{-1}\mbs{\mu}_1 + \mbf{W}_{i2}\mbs{\Sigma}_2^{-1}\mbs{\mu}_2\right),
    \label{eq:merge_sigma}
\end{align}
where $\mbs{\est{i}{k}} \sim \mc{N}(\tilde{\mbs{\mu}}_i, \tilde{\mbs{\Sigma}}_i)$ where $i \in \{1,2\}$. Hence, the beliefs obtained after the consensus step~\eqref{eq:estimate} remain Gaussian implying the corresponding predictive distribution also remains Gaussian.

\begin{figure}[!htb]

\centering
\subfloat[]{\label{main:a}\includegraphics[width=0.8\columnwidth]{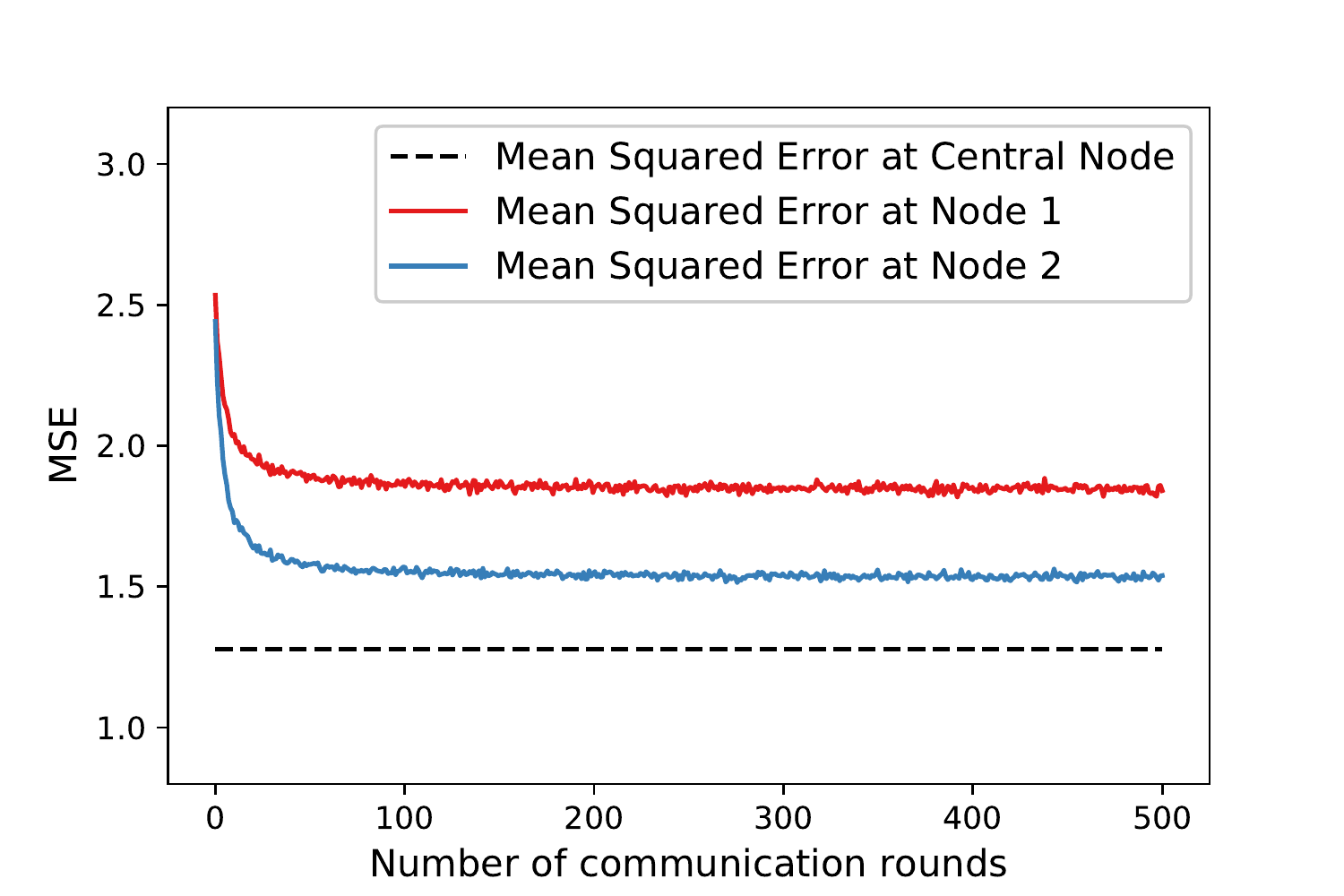}}

\centering
\subfloat[]{\label{main:b}\includegraphics[width=0.8\columnwidth]{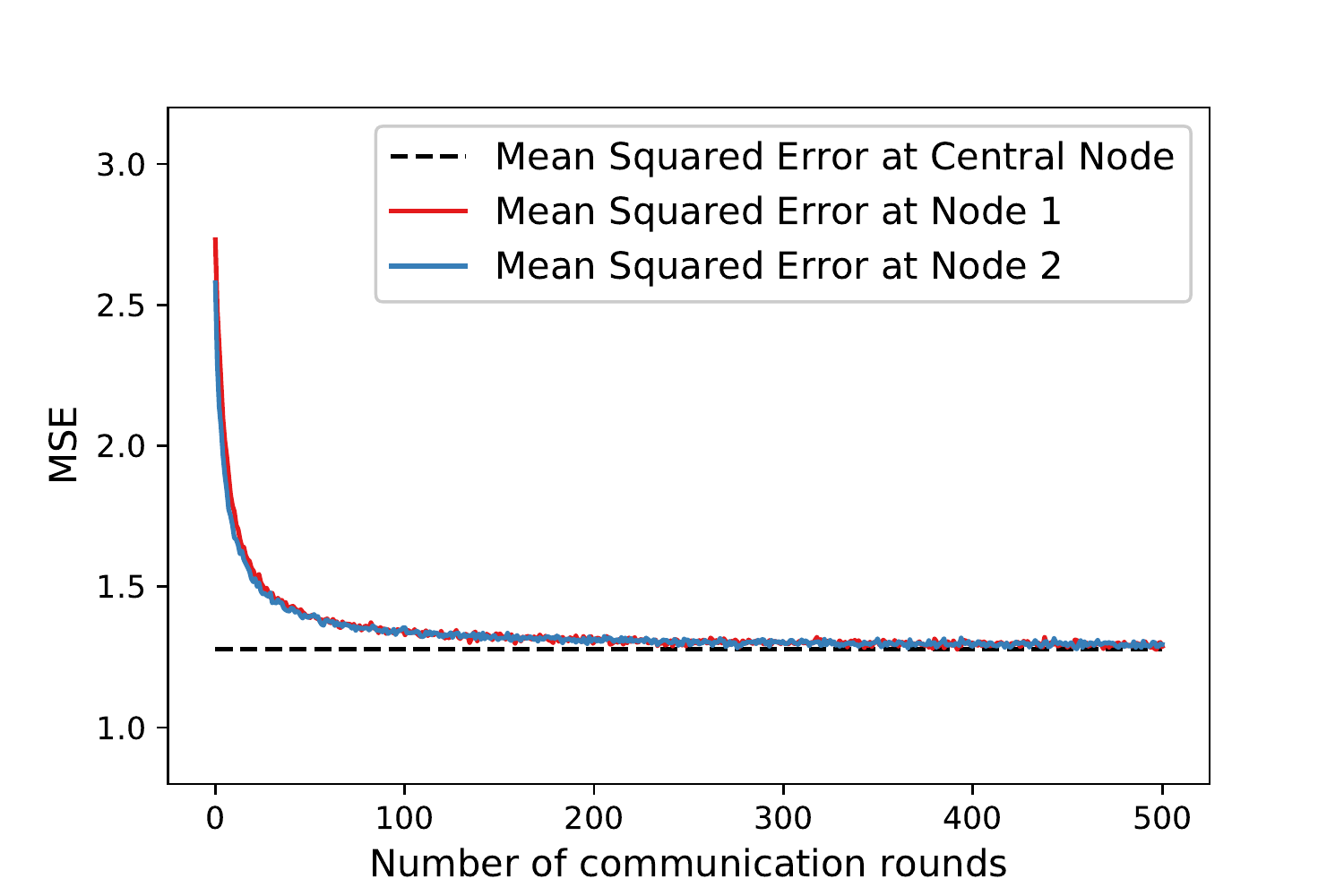}}

\caption{Figure shows the MSE of Bayesian prediction on the test set samples over time for two cases: (a) node 1 and 2 trained with no cooperation (b) node 1 and 2 trained using the proposed learning rule.}\vspace{-0.2cm}
\label{fig:main}
\end{figure}

Now we compare the mean squared error (MSE) of the predictions over the test set, when nodes are trained using Algorithm~\ref{alg:algorithm}, with that of two cases: (1) a central node which has access to training data samples $\mbf{x} =[x_1, x_2]^T \in \mbb{R}$ where $x_1$ is sampled from $\text{Unif}[-1,1]$ and  $x_2$ is sampled from $\text{Unif}[-1.5, 1.5]$, and (2) nodes learn without without cooperation using local training data only. Figure~\ref{fig:main}(a) shows that the MSE of both nodes, when trained without cooperation, is higher than that of central node implying the performance of nodes has degraded due to insufficient local information to learn $\mbs{\theta}^{\ast}$. However, Figure~\ref{fig:main}(b) shows that the MSE of both nodes, when trained using Algorithm~\ref{alg:algorithm}, matches that of a central node implying that the nodes were able to correctly learn true $\mbs{\theta}^{\ast}$.

    
    

\begin{remarks}
Note that Gaussian likelihood functions considered in Example~\ref{ex:dist_linear_regression} violate the bounded likelihood functions assumption. Furthermore, the parameters belong to a continuous parameter set $\Theta$. This example and those that follow demonstrate that our analytical assumptions on the likelihood functions and the parameter set are sufficient but not necessary for convergence of our distributed learning rule.
\end{remarks}

\vspace{-0.2cm}
\subsection{Distributed Training of DNN Models}
\subsubsection{Training 
DNN Models}
\label{subsec:practical}

Each round of the Algorithm~\ref{alg:algorithm} involves a local Bayesian update (\ref{eq:bayes}), followed by a consensus step (\ref{eq:estimate}). For most practical problems, the exact computation of the normalizing  constants in these update rules is computationally intractable. In this section, we propose modifications to Algorithm~\ref{alg:algorithm}, to make it more suitable for learning DNN models. In particular, we consider Variational Inference (VI) \citep[Chapter~3]{gal2016uncertainty} techniques which replace the Bayesian modelling marginalization with optimization. 
    
    \underline{Modification to Bayesian Update~\eqref{eq:bayes}:} 
    Let $q_{\varphi} \in \mc{P}(\Theta)$ denote an approximating variational distribution, parametrized by $\varphi$, which is  easy to evaluate such as the exponential family. We want to find an approximating distribution which is as close as possible to the posterior distribution obtained using equation~\eqref{eq:bayes}. In other words, given a prior $\est{i}{k}(\theta)$ for all $\theta \in \Theta$ and the likelihood functions $\{l_i(y; \theta, x): y \in \mc{Y}, \, \theta \in \Theta, \, x \in \mc{X}_i\}$ we want to learn an approximate posterior $q_{\varphi}(\cdot)$ over $\Theta$ at each node. This involves maximizing the evidence lower bound (ELBO) with respect to the variational parameters defining $q_{\varphi}(\cdot)$,
    \begin{align}
    \label{eq:variational}
    \mc{L}_{\text{VI}}(\varphi) 
    :=&  -\int_{\Theta} q_{\varphi}(\theta) \log \dist{i}{y}{\theta, x} d\theta 
    \nonumber
    \\
    &\hspace{1cm}+ D_{\text{KL}}\left(q_{\varphi}(\theta) || \est{i}{k}(\theta) \right) .
    \end{align}
    Furthermore, instead of performing updates after every observed training sample, a batch of observations can be used for obtaining the approximate posterior update using VI techniques.
    
    \underline{Modification to Consensus Step~\eqref{eq:estimate}:}
    Similarly, the consensus step given by equation~\eqref{eq:estimate} can also be computationally intractable in practical applications due to the normalization involved. Hence, we propose the use of an unnormalized belief vector $\mbs{\est{i}{k}}$ and claim that this does not alter the optimization problem in equation~\eqref{eq:variational}. More specifically, for any $\kappa>0$, we have
    \begin{align*}
    D_{\text{KL}}\left( q_{\varphi}(\theta) || \kappa \est{i}{k}(\theta)\right) 
    = D_{\text{KL}}\left( q_{\varphi}(\theta) || \est{i}{k}(\theta)\right)  - \log \kappa .
    \end{align*} 
    Thus we can perform the consensus update with the unnormalized beliefs.

\subsubsection{Classification Fashion MNIST}

We consider the problem of distributed training of two nodes for classifying the MNIST fashion dataset \cite{xiao2017_online}. At each node we train a fully connected neural network, with one hidden layer consisting of 400 units. This dataset consists of 60,000 training pixel images and 10,000 testing pixel images of size 28 by 28. Each image is labelled with its corresponding number (between zero and nine, inclusive). Let $\mc{D}_i$ for $i \in \{1,2\}$ denote the local training dataset at each node $i$. We aim to train each neural network to learn a distribution over its weights $\mbs{\theta}$, i.e., the posterior distribution $\P(\mbs{\theta}|\mc{D}_i)$ at each node $i$.

When the nodes are training without cooperation we directly employ VI techniques to learn a Gaussian variational approximation to $\P(\mbs{\theta}|\mc{D}_i)$. More specifically, we choose the approximating family of distributions to be Gaussian distributions which belong to the class $\{q(\cdot; \mbs{\mu}, \mbs{\Sigma}): \mu \in \mbb{R}^d, \, \mbs{\Sigma}= \text{diag}(\mbs{\sigma}), \, \mbs{\sigma} \in \mbb{R}^d\}$, where $d$ is the number of weights in the neural network.  We use the Bayes by Backprop training algorithm proposed by~\citet{weight_uncert_NN} to learn the Gaussian variational posterior. We then sample weights from the variational posterior to make predictions on the test set. Next, we embed the nodes in an aperiodic network with edge weights given by $\mbf{W}$. When the nodes are training with cooperation over this network, they employ Algorithm~\ref{alg:algorithm} with variational inference in place of the Bayesian update step. Again we use Bayes by Backprop training algorithm to learn the Gaussian variational posterior at each node. Furthermore, since the approximating distributions for $\mbs{\bel{i}{k}}$ are Gaussian distributions, the consensus step reduces to equations~\eqref{eq:merge_mu} and~\eqref{eq:merge_sigma}, and we obtain $\mbs{\est{i}{k}} \sim \mc{N}(\tilde{\mbs{\mu}}_i, \tilde{\mbs{\Sigma}}_i)$ for $i \in \{1,2\}$.  

For the architecture we have considered, a central node with access to all the training samples obtains an accuracy of $88.28\%$. First, we divide fashion MNIST training set in an i.i.d manner where each $\mc{D}_i$ consists of half of the training set samples. In this setting, accuracy at node 1 is $87.07\%$ without cooperation and $87.43\%$ with cooperation, and at node 2 it is $87.43\%$ without cooperation and $87.84\%$ with cooperation. We observe that there is no loss in accuracy due to cooperation. Next, we consider two more interesting and challenging settings of distributed training: 

    \textbf{Non-IID and Balanced:} Data at each node is obtained using different labelling distributions. We consider two cases of this setting: case (a) $\mc{D}_1$ consists of training samples with labels only in $\{0,1,2,3,4\}$ and $\mc{D}_2$ consists of training samples with labels only in $\{5,6,7,8,9\}$; and case (b) $\mc{D}_1$ consists of training samples with labels only in classes $\{0,2,3,4,6\}$ and $\mc{D}_2$ consists of training samples with labels only in $\{1,5,7,8,9\}$. When the nodes cooperate we consider a weight matrix $\mbf{W} =\begin{bmatrix} 0.25 & 0.75\\ 0.75 & 0.25 \end{bmatrix}$.
    
    \textbf{Non-IID and Unbalanced:} Number of training samples at each node is highly unbalanced. We consider the case where $\mc{D}_1$ consists of training samples with labels only in $\{0,1,2,3,4,5,6,7\}$ and $\mc{D}_2$ consists of training samples with labels only in $\{8,9\}$. When the nodes cooperate we consider a weight matrix $\mbf{W} =\begin{bmatrix} 0.45 & 0.55\\ 0.70 & 0.30 \end{bmatrix}$.

\begin{figure*}[!bth]
\begin{minipage}{\columnwidth}
\centering
\subfloat[]{\label{balanced_acc:a}\includegraphics[width=0.8\columnwidth, trim={1mm 1mm 8mm 1mm},clip]{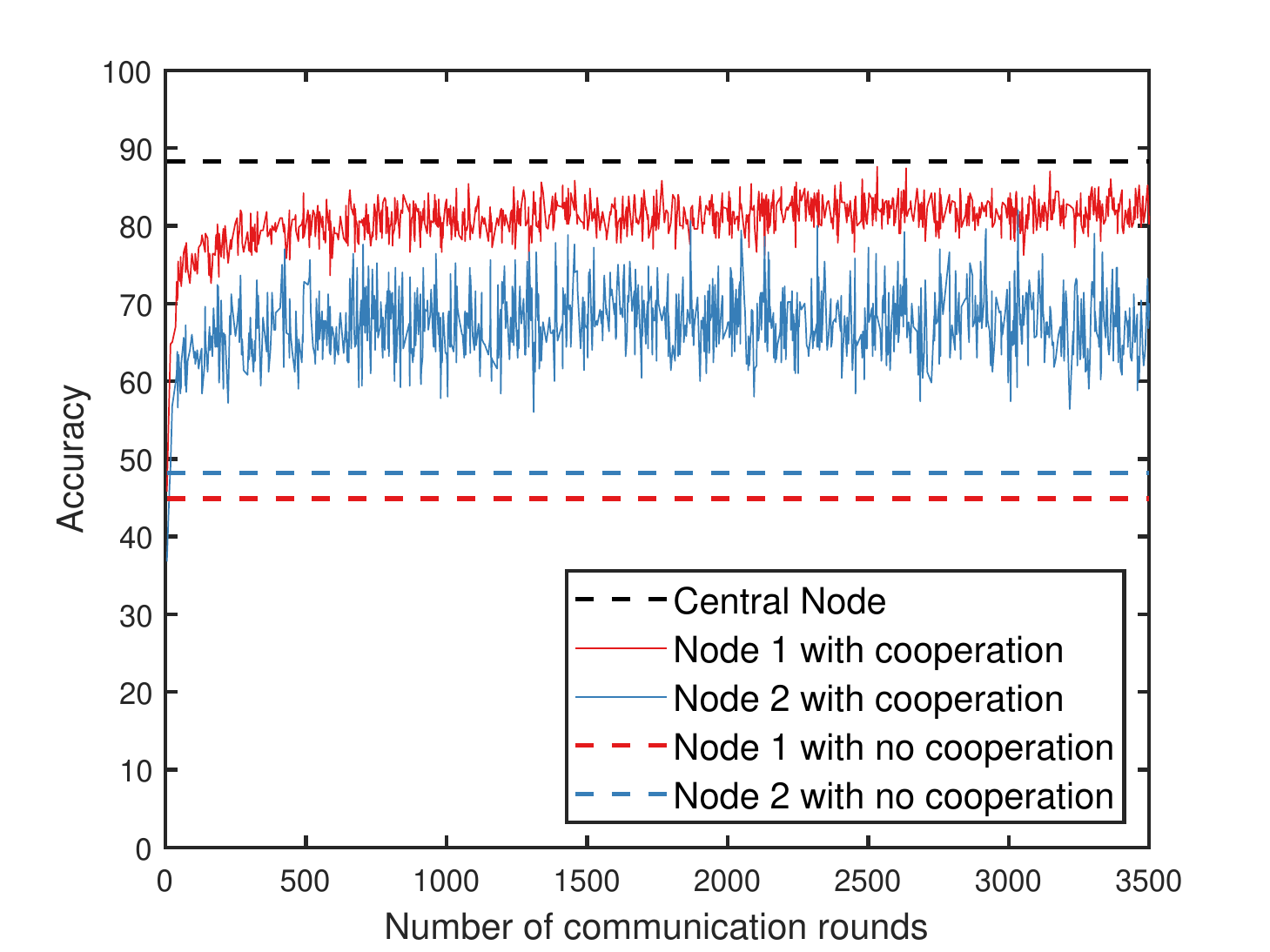}}
\end{minipage}%
\begin{minipage}{\columnwidth}
\centering
\subfloat[]{\label{balanced_acc:b}\includegraphics[width=0.8\columnwidth, trim={1mm 1mm 8mm 1mm},clip]{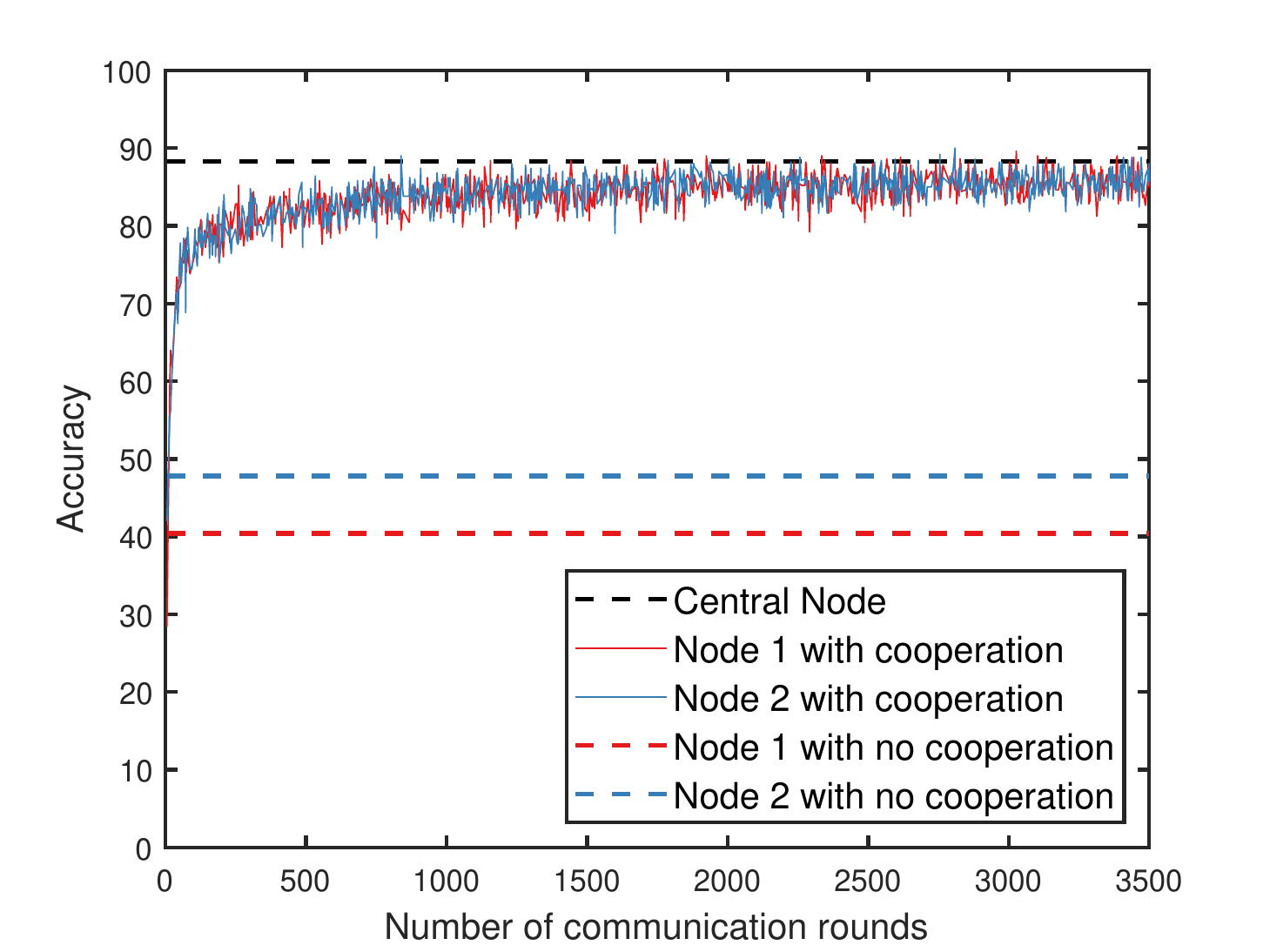}}
\end{minipage}
\caption{Figure shows the accuracy obtained for distributed training under the setting of non-IID balanced case (a) on the left and case (b) on the right.} 
\end{figure*}

\begin{figure*}[!bth]

\begin{minipage}{.5\linewidth}
\centering
\subfloat[]{\label{balanced:a}\includegraphics[width=0.8\columnwidth]{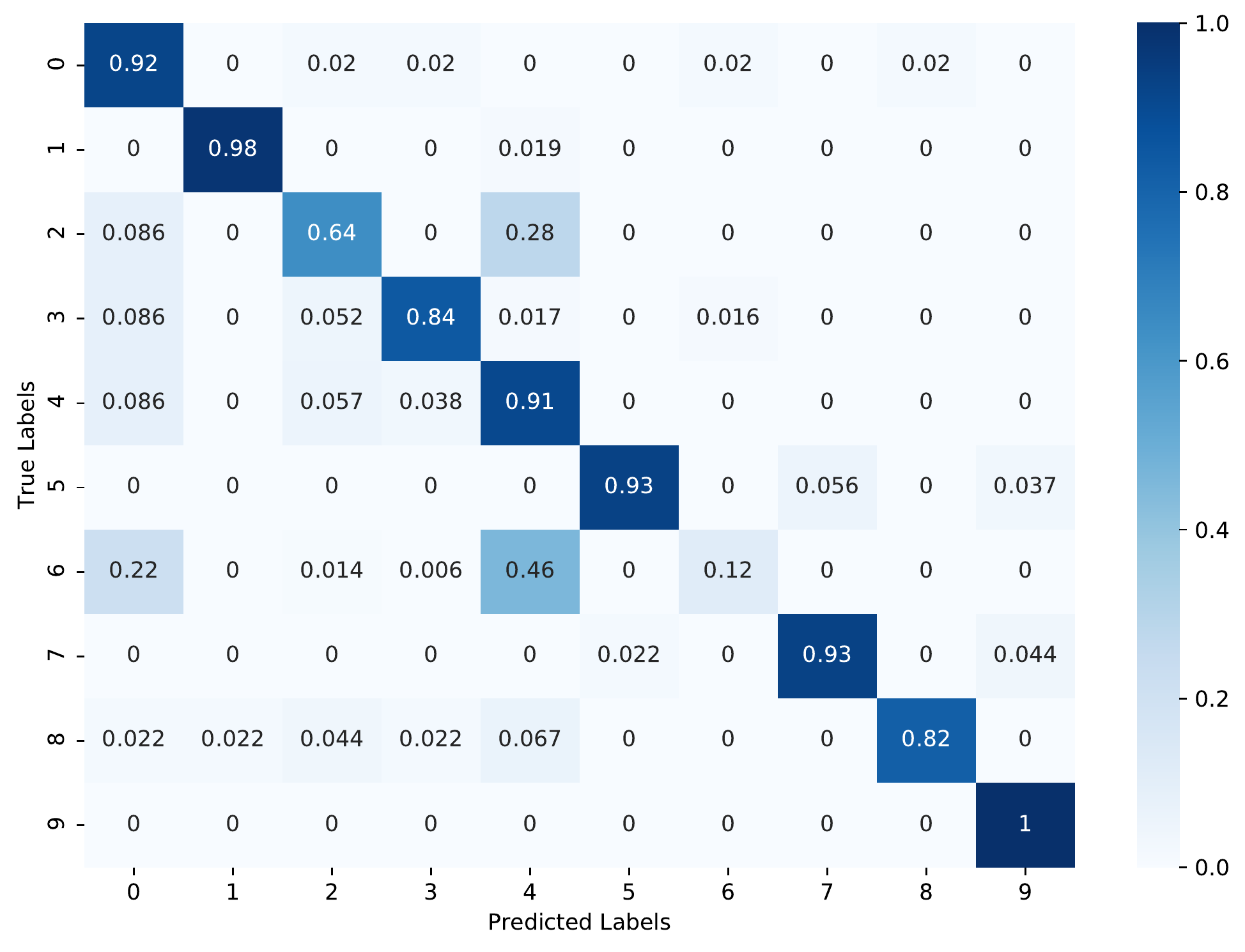}}
\end{minipage}%
\begin{minipage}{.5\linewidth}
\centering
\subfloat[]{\label{balanced:b}\includegraphics[width=0.8\columnwidth]{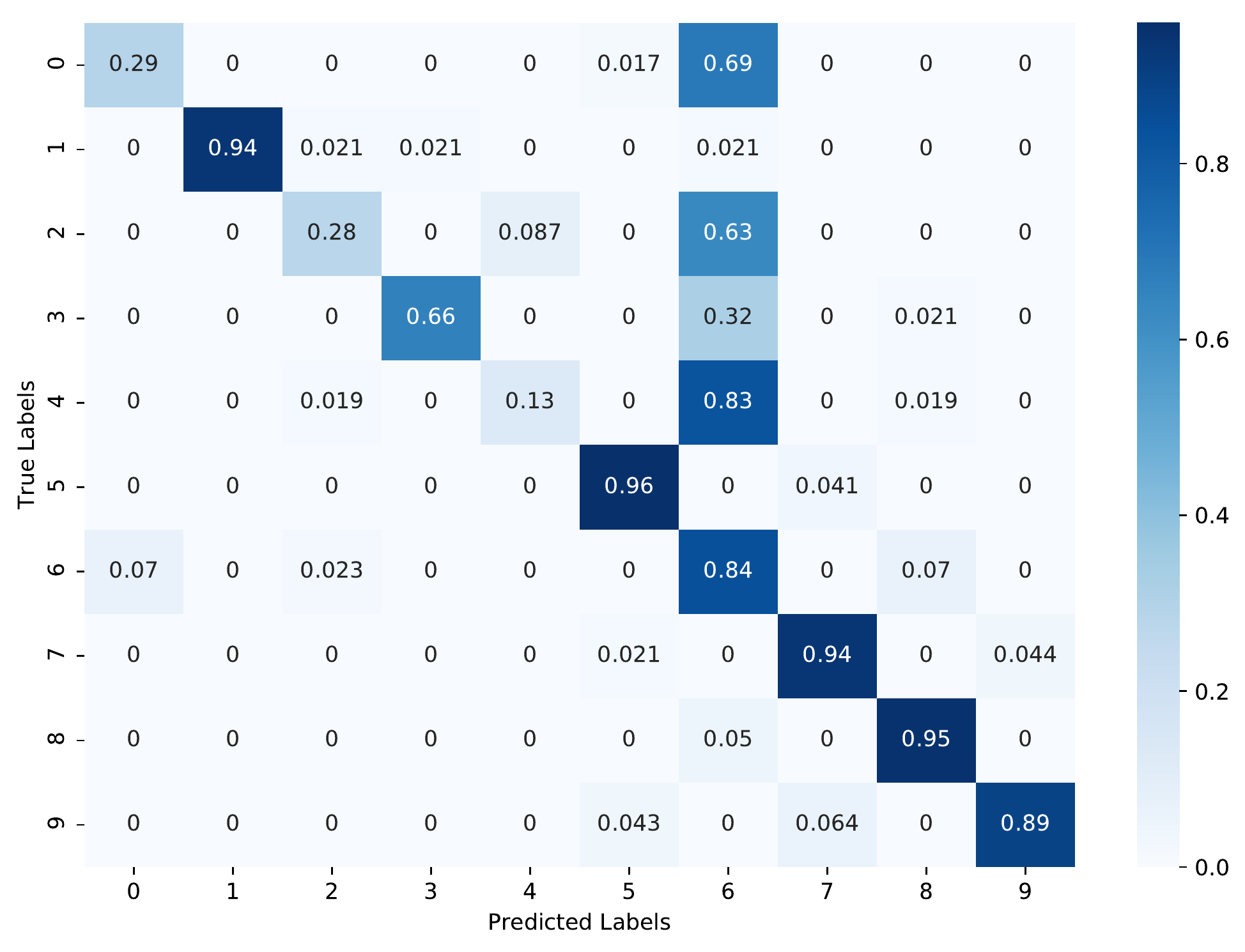}}
\end{minipage}

\begin{minipage}{.5\linewidth}
\centering
\subfloat[]{\label{balanced:c}\includegraphics[width=0.8\columnwidth]{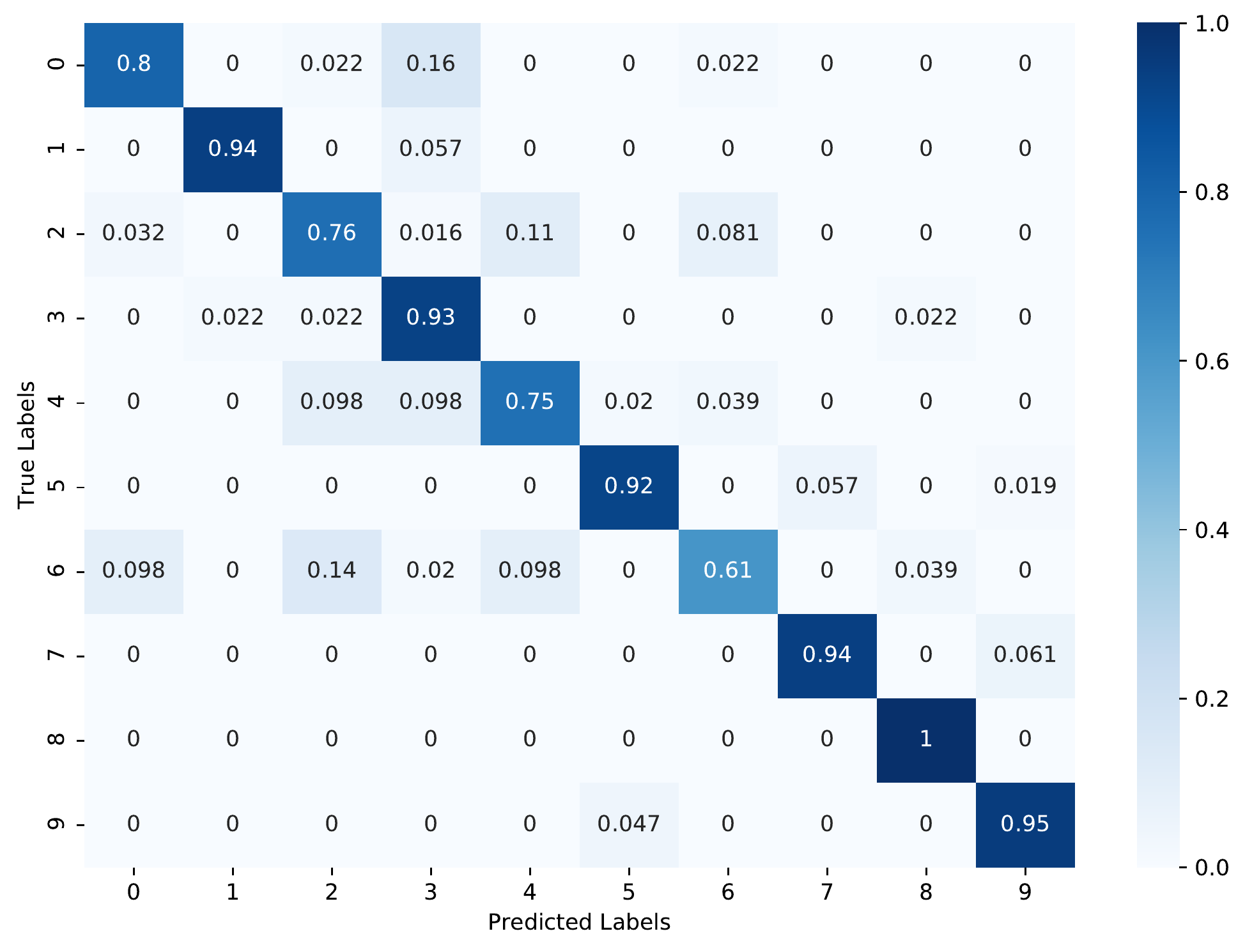}}
\end{minipage}%
\begin{minipage}{.5\linewidth}
\centering
\subfloat[]{\label{balanced:d}\includegraphics[width=0.8\columnwidth]{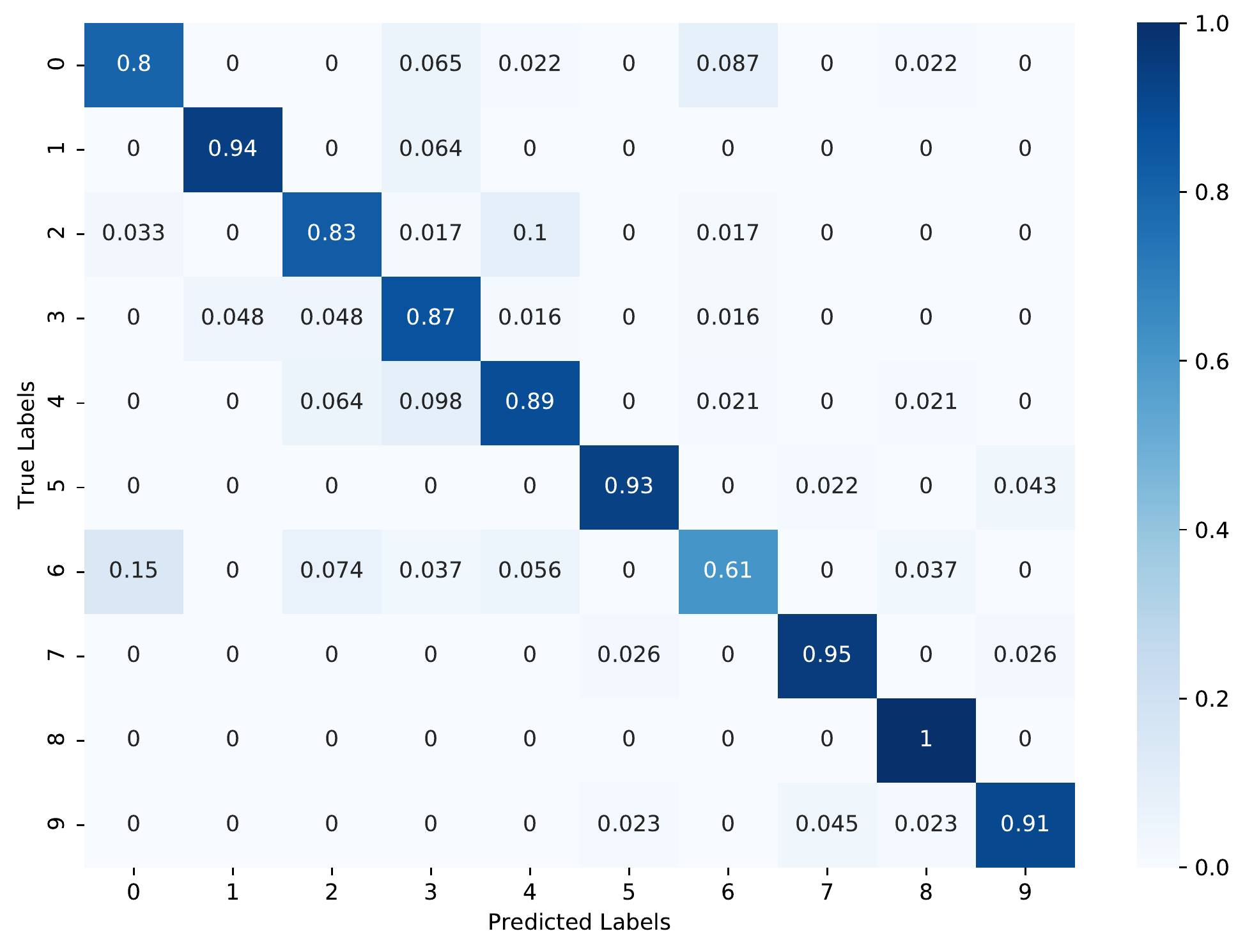}}
\end{minipage}

\caption{Figure shows the confusion matrices for node 1 on the left and node 2 on the right, under the setting of non-IID balanced case (a) in the top row and case (b) in the bottom row.}\vspace{-0.2cm}
\end{figure*}

In the setting of non-IID and balanced case (a), when the nodes train without cooperation, node 1 and node 2 obtain an accuracy of $44.89\%$ and $48.22\%$ respectively. However, when using Algorithm~\ref{alg:algorithm}, the accuracy improves to $83\%$ and $67\%$ respectively as shown in Figure~\ref{balanced_acc:a}. Now we will examine the accuracy of labels $\{0,2,3,4,6\}$ since this set labels corresponds to the labels: t-shirt (0), pullover (2), dress (3), coat (4), and shirt (6); which look similar to each other compared to other labels. From Figure~\ref{balanced:a} we observe that since node 1 has access to training samples for the classes $\{0,2,3,4\}$ except class 6, it misclassifies class 6 as $\{0,2,3,4\}$ whereas other classes including those inaccessible to node 1 get classified correctly with high accuracy. Similarly, from Figure~\ref{balanced:b} we observe that since node 2 has access to training samples for the class 6 but not for classes $\{0,2,3,4\}$, these classes often get misclassified as class 6. This explains the poor accuracy obtained at node 2 compared to the accuracy obtained at node 1. 

In the setting of non-IID and balanced case (b), when the nodes train without cooperation node 1 and node 2 obtain an accuracy of $40.4\%$ and $47.8\%$ respectively and with cooperation nodes obtain an accuracy of $85.78\%$ and $85.86\%$ respectively. Again we will examine the accuracy of labels $\{0,2,3,4,6\}$. From Figure~\ref{balanced:c} we observe that since node 1 has access to training samples for the classes $\{0,2,3,4,6\}$ it obtains high accuracy in those classes. As shown Figure~\ref{balanced:d} since node 2 is learning from the expert node 1 it no longer misclassifies the classes $\{0,2,3,4,6\}$. Hence, in this setup both nodes obtain high accuracy. We conclude that a setup where each node is an expert at its local task distributed training turns every node into an expert on the network-wide task.

\begin{figure}[!htb]
\centering
    \includegraphics[width=0.8\columnwidth, trim={1mm 1mm 8mm 1mm},clip]{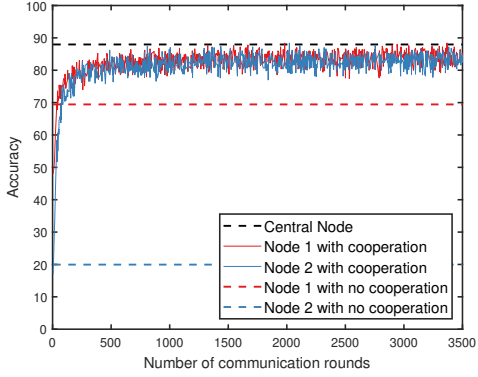}
   \caption{Figure shows the accuracy obtained for distributed training under the setting of non-IID unbalanced case.}
   \label{fig:non_iid_unbal}
\end{figure}

In the setting of non-IID and unbalanced case, when nodes train without cooperation node 1 and node 2 obtain an accuracy of $69.44\%$ and $19.95\%$ respectively and with cooperation nodes obtain an accuracy of $85.8\%$ and $85.2\%$ respectively. Figure~\ref{fig:non_iid_unbal} shows that a single export at a network-wide task can pull up the accuracy of other nodes.

\vspace{-0.2cm}
\section{Discussion and Future Work}
\label{sec:discussion}
In this paper, we considered the problem of decentralized learning over a network of nodes with no central server. We considered a peer-to-peer learning algorithm in which nodes iterate and aggregate the beliefs of their one-hop neighbors and collaboratively estimate the global optimal parameter. We obtained high probability bounds on the network wide worst case probability of error, and also discussed suitable approximations for applying this algorithm for learning DNN models. Our experiments results show encouraging results with negligible drop in accuracy when training happens in a decentralized manner. An important area of future work includes extensive empirical studies with various DNN architectures. An interesting area of future work is apply this learning algorithm to random graphs~\cite{7172262} and to decentralized reinforcement learning~\cite{8278014} 

\vspace{-0.2cm}
\appendix
\section{Proof of Theorem~1}
\label{app:proof}
The proof of Theorem~1 is based the proof provided by~\citet{7172262, 7349151, 8359193}. For the ease of exposition, let $\est{i}{0}(\theta) =\frac{1}{M}$ for all $\theta \in \Theta$. Consider some $\theta^{\ast} \in \Theta^{\ast}$. We begin with the following recursion for each node $i \in [N]$ and for any $\theta \not\in \Theta^{\ast}$, \vspace{-0.2cm}
\begin{align*}
 & \frac{1}{n}\log \frac{\est{i}{n}(\theta^{\ast})}{\est{i}{n}(\theta)} 
= \frac{1}{n}\sum_{j = 1}^{N}\sum_{k = 1}^{n} W^{k}_{ij}z_j^{(n-k+1)}(\theta^{\ast}, \theta),
 \end{align*}
 where\vspace{-0.2cm}
 \begin{align*}
z_j^{(k)}(\theta^{\ast}, \theta) = \log \frac{\dist{j}{\samp{j}{k}}{\theta^{\ast}, X_i^{(k)}} }{\dist{j}{\samp{j}{k}}{\theta, X_i^{(k)}}}.
 \end{align*}
\vspace{-0.2cm} From the above recursion we have
\begin{align*}
&\frac{1}{n}\log \frac{\est{i}{n}(\theta^{\ast})}{\est{i}{n}(\theta)}
\geq 
\frac{1}{n}\sum_{j = 1}^{N} v_j\left(\sum_{k = 1}^{n}  z_j^{(k)}(\theta^{\ast}, \theta)\right)
\\
& \hspace{1cm}-\frac{1}{n}\sum_{j = 1}^{N} \sum_{k = 1}^{n} \left| W^{k}_{ij} - v_j \right| \left| z_j^{(k)}(\theta^{\ast}, \theta) \right|
\\
& \overset{(a)}\geq
\frac{1}{n}\sum_{j = 1}^{N} v_j \left(\sum_{k = 1}^{n} z_j^{(k)}(\theta^{\ast}, \theta)\right)-\frac{4C\log N}{n(1-\lambda_{\text{max}}(W))},
\end{align*}
where $(a)$ follows from Lemma~\ref{lemma:conv_W} and the boundedness assumption of log-likelihood ratios. Now fix $n \geq \frac{8C\log N}{\epsilon (1-\lambda_{\text{max}}(W))}$, since $\est{i}{n}(\theta^{\ast}) \leq 1$ we have
\begin{align*}
     -\frac{1}{n}\log \est{i}{n}(\theta) 
     \geq
     -\frac{\epsilon}{2}
     +
     \frac{1}{n}\sum_{j = 1}^{N} v_j \left(\sum_{k = 1}^{n} z_j^{(k)}(\theta^{\ast}, \theta)\right).
\end{align*}
\vspace{-0.1cm}Furthermore, we have\vspace{-0.2cm}
\begin{align*}
    &\P\left( -\frac{1}{n}\log \est{i}{n}(\theta)  \leq \sum_{j = 1}^{N} v_j I_j(\theta^{\ast}, \theta) - \epsilon \right)
    \\
    &\leq 
    \P\left( 
     \frac{1}{n}\sum_{j = 1}^{N} v_j \sum_{k = 1}^{n} z_j^{(k)}(\theta^{\ast}, \theta)  \leq  \sum_{j = 1}^{N} v_j I_j(\theta^{\ast}, \theta) - \frac{\epsilon}{2} \right).
\end{align*}

Now for any $j \in [N]$ note that\vspace{-0.2cm}
\begin{align*}
   &\sum_{j = 1}^{N} v_j\sum_{k = 1}^{n}   z_j^{(k)}(\theta^{\ast}, \theta) 
   - n \sum_{j = 1}^{N} v_j I_j(\theta^{\ast}, \theta)
   \\ 
   &= \sum_{k = 1}^{n}   \left( \sum_{j = 1}^{N} v_jz_j^{(k)}(\theta^{\ast}, \theta) - \sum_{j = 1}^{N} v_j\expe[z_j^{(k)}(\theta^{\ast}, \theta)]\right).
\end{align*}
For any $\theta \not\in \Theta^{\ast}$, applying McDiarmid's inequality for all $\epsilon > 0$  and for all $n \geq 1$ we have\vspace{-0.2cm}
\begin{align*}
    &\P\left( \sum_{k = 1}^{n}   \left( \sum_{j = 1}^{N} v_j z_j^{(k)}(\theta^{\ast}, \theta) - \sum_{j = 1}^{N} v_j\expe[z_j^{(k)}(\theta^{\ast}, \theta)]\right) \right. 
    \\
    &\left. \hspace{2cm}\leq -\frac{\epsilon n}{2}  \right)
    \leq 
    e^{-\frac{\epsilon^2 n}{2C}}.
\end{align*}

\vspace{-0.2cm}
Hence, for all $\theta \not\in \Theta^{\ast}$, for $n \geq \frac{8C\log N}{\epsilon (1-\lambda_{\text{max}}(W))}$ we have\vspace{-0.2cm}
\begin{align*}
    \P\left(\frac{-1}{n}\log \est{i}{n}(\theta)  \leq \sum_{j = 1}^{N} v_jI_j(\theta^{\ast}, \theta) - \epsilon\right)  \leq 
    e^{-\frac{\epsilon^2 n}{4C}},
\end{align*}
which implies\vspace{-0.2cm}
\begin{align*}
    \P\left(\est{i}{n}(\theta) \geq e^{-n( \sum_{j = 1}^{N} v_jI_j(\theta^{\ast}, \theta) - \epsilon)} \right)  \leq 
    e^{-\frac{\epsilon^2 n}{4C}}.
\end{align*}
Using this we obtain a bound on the worst case error over all $\theta$ and across the entire network as follows\vspace{-0.2cm}
\begin{align*}
    \P\left(\max_{i \in [N]}\max_{\theta \not\in \Theta^{\ast}}\est{i}{n}(\theta) \geq e^{-n( K(\Theta) - \epsilon)} \right)  \leq 
    N Me^{-\frac{\epsilon^2 n}{4C}},
\end{align*}
where $K(\Theta):= \min_{\theta \in \Theta^{\ast}, \psi \in \Theta\setminus \Theta^{\ast}}\sum_{j = 1}^{N} v_j I_j(\theta, \psi)$. From Assumption~\ref{assump:global_learnability} and Lemma~\ref{lemma:conv_W} we have that $K(\Theta) >0$. Choose $\epsilon = \frac{K(\Theta)}{2}$. Therefore,  for a given confidence parameter $\delta \in (0,1)$ we have 
\begin{align*}
    &\P\left(\exists \, i \in [N] \,\, \text{s.t.} \,\, \hat{\theta}^{(n)}_i \not \in \Theta^{\ast} \right)
    \\
    & \leq \P\left(\max_{i \in [N]}\max_{\theta \not\in \Theta^{\ast}}\est{i}{n}(\theta) \geq \frac{1}{2 M} \right)  \leq 
    \delta,
\end{align*}
when the number of samples satisfies
\begin{align*}
n 
\geq 
\frac{16C\log \frac{N M}{\delta}}{ K(\Theta)^2(1-\lambda_{\text{max}}(W))}.
\end{align*}

\begin{lemma}[~\citet{7349151}]
\label{lemma:conv_W}
For an irreducible and aperiodic $W$, the stationary distribution $\mbf{v} = [v_1, v_2, \ldots, v_N]$ is unique and has strictly positive components and satisfies $v_i = \sum_{j = 1}^{n}v_j W_{ji}$. Furthermore, for any $i \in [N]$ the weight matrix satisfies\vspace{-0.2cm}
\begin{align*}
    \sum_{k = 1}^n \sum_{j=1}^N \left| W^{k}_{ij} - v_j \right| \leq \frac{4\log N}{1-\lambda_{\text{max}}(W)},
\end{align*}
where $\lambda_{\text{max}}(W) = \max_{i \in [N-1]}\lambda_i(W)$, and $\lambda_i(W)$ denotes eigenvalue of $W$ counted with algebraic multiplicity and $\lambda_0(W) = 1$.
\end{lemma}

\clearpage

\bibliography{ref}
\bibliographystyle{icml2019}



\end{document}